# Aligned Manifold Property and Topology Point Clouds for Learning Molecular Properties

An Efficient Molecular Surface Representation Integrating Local Quantum Properties and Topological Context Without Requiring SE(3)-Invariant Architectures

Alexander Mihalcea[1]  –   contact email: amihalce@purdue.edu;                                         July 21, 2025

Machine learning models for molecular property prediction generally rely on representations—such as SMILES strings and molecular graphs—that overlook the surface-local phenomena driving intermolecular behavior. 3D-based approaches often reduce surface detail or require computationally expensive SE(3)-equivariant architectures to manage spatial variance. To overcome these limitations, this work introduces AMPTCR (Aligned Manifold Property and Topology Cloud Representation), a molecular surface representation that combines local quantum-derived scalar fields and custom topological descriptors within an aligned point cloud format. Each surface point includes a chemically meaningful scalar, geodesically derived topology vectors, and coordinates transformed into a canonical reference frame, enabling efficient learning with conventional SE(3)-sensitive architectures. AMPTCR is evaluated using a DGCNN framework on two tasks: molecular weight and bacterial growth inhibition. For molecular weight, results confirm that AMPTCR encodes physically meaningful data, with a validation $R^2$ of 0.87. In the bacterial inhibition task, AMPTCR enables both classification and direct regression of E. coli inhibition values using Dual Fukui functions as the electronic descriptor and Morgan Fingerprints as auxiliary data, achieving an ROC AUC of 0.912 on the classification task, and an $R^2$ of 0.54 on the regression task. These results help demonstrate that AMPTCR offers a compact, expressive, and architecture-agnostic representation for modeling surface-mediated molecular properties.

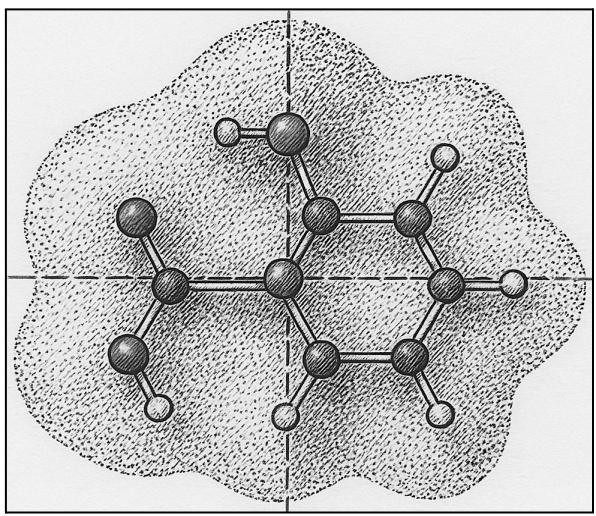
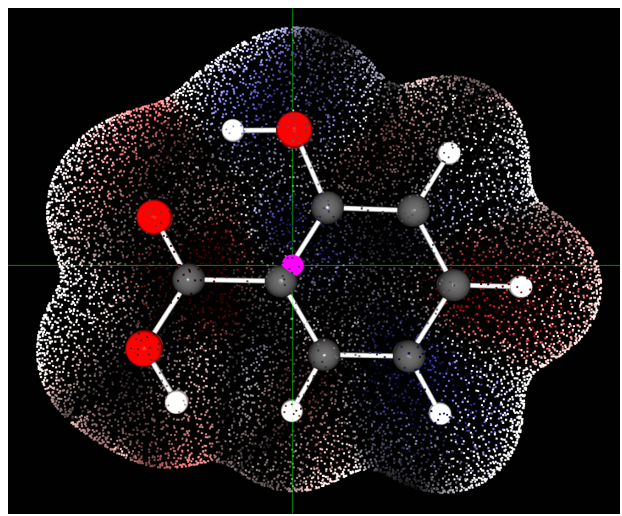

---

[1] This author is an undergraduate student at Purdue University majoring in Pharmaceutical Sciences along with a minor in Machine Learning. This work was conducted entirely independently and is not affiliated with, sponsored by, or endorsed by the university. All intellectual property rights are retained solely by the author.

# Introduction

Molecular property prediction has recently fallen into the purview of machine learning, with notable success across multiple domains. However, a core aspect that limits what a neural network will be capable of learning is how the input is represented. In the case of molecules, these representations often ignore the fundamental mechanism by which they interact—with their electron surfaces, not just their atoms. SMILES strings and graphs, the most common molecular representations, were never designed to capture surface-local electronic properties or geometric features relevant to much of intermolecular behavior. Even 3D-based methods often fall short, either by reducing surfaces to coarse voxel grids or by offloading spatial awareness to computationally expensive SE(3)-equivariant architectures. In the process, the structure–function relationship encoded at the nanoscale surface level is lost.

For numerous properties, including—but not limited to—solubility, minimum inhibitory concentration (MIC), binding affinity, and chemical reactivity, surface-local phenomena often play a dominant role. These effects are influenced not only by the molecule's global conformation but also by fine-grained variations in surface curvature, quantum-derived properties such as electrostatic potential (ESP) or Fukui functions, and the surrounding topological structure. Accurately capturing these features requires a representation that is sensitive to local geometry and electronic context, while remaining robust to the non-deterministic orientation of molecular structures.

Current solutions frequently address this challenge at the architectural level, introducing additional model complexity, computational overhead, and constraints on generalizability. Other methods attempt to project the 3D surface manifold onto a 2D embedding ([e.g., Li et al.](#)), which can significantly compromise the fidelity of topological and geometric information. This information loss is particularly limiting for tasks where the precise shape and features of the molecular surface are tightly coupled to functional behavior—something especially prevalent in biological systems, where surface-mediated interactions often determine molecular efficacy and selectivity.

To address these limitations, this work introduces **AMPTCR**: an **Aligned Manifold Property and Topology Cloud Representation** for learning the properties of small molecules, especially those of interest in pharmaceutical drug development. AMPTCR encodes each surface point as a compact feature vector that captures local quantum-derived scalar fields and intrinsic surface topology, projected into a canonical reference frame using a custom alignment heuristic. This alignment enables direct use with lightweight, SE(3)-sensitive architectures without relying on explicit rotational invariance at the model level.

This preprint outlines the motivation and design rationale behind AMPTCR and evaluates its performance on molecular property prediction tasks where surface interactions are expected to be critical, with bacterial growth inhibition serving as the primary evaluation endpoint. Due to ongoing intellectual property protection, certain technical details are intentionally withheld; however, the core representation framework and its utility are described and benchmarked in sufficient detail to support performance and future exploration.

# AMPTCR Representation

AMPTCR (Aligned Manifold Property and Topology Cloud Representation) encodes the molecular surface as a structured point cloud, with each surface point annotated by a local quantum-derived scalar and a compact set of geometric-topological features. The representation is designed to be lightweight, architecture-agnostic, and spatially aware, allowing models to reason about molecular surfaces without relying on SE(3)-equivariant architectures.

### Surface Construction and Scalar Annotation

Molecular surfaces are generated from .PDB files using quantum chemical calculations to extract local electron density information. A composite scalar field is derived from charge-perturbed density grids and used to color the surface, producing a mesh representation that reflects local chemical reactivity. The surface is then sampled to a fixed number of evenly distributed points. The scalar field at each point reflects chemically meaningful surface properties—such as electrostatic potential (ESP) or Dual Fukui functions—and is normalized per-molecule. The resulting scalar is stored as a single floating-point value per surface point. Mesh generation and processing leverage a combination of Python tools, including Psi4 for quantum calculations, scikit-image for surface extraction, and Trimesh for mesh manipulation.

### Intrinsic Alignment Heuristic

To avoid the architectural burden of enforcing rotation and translation invariance, each molecule is aligned into a consistent local reference frame prior to learning. The alignment is derived from both molecular geometry and the spatial distribution of the surface scalar field. This procedure is meant to be deterministic, fast, and preserves local spatial relationships across molecules in a chemically meaningful way. The alignment is particularly important in the context of 3D learning. While enforcing invariance in 1D or 2D typically adds minimal overhead, fully equivariant 3D models can slow training by anywhere from 10 to 30 times ([Liu et al.](#)). AMPTCR shifts this burden from the model to the data, enabling fast training and compatibility with standard architectures. The exact methods used are not described within this preprint.

### Topological Encoding

Each surface point is further embedded with a set of "topology vectors" that encode local curvature and directional structure within the surface manifold. These vectors are derived from the geometry of neighboring regions, sampled through an intrinsic geodesic framework. The resulting descriptors are compact, rotation-sensitive, and engineered to capture subtle geometric cues relevant to molecular function—such as whether a point lies at the apex of a hill or the bottom of a valley, as well as the magnitude of surrounding curvature. This topological embedding fills a critical gap left by conventional point cloud representations, where sparsely sampled points can make it difficult for neural networks to infer detailed surface topology on their own. By explicitly encoding topological information at each point, AMPTCR provides neural networks with direct access to rich geometric context, enabling more accurate learning of how surface curvature affects molecular properties. The exact construction of these descriptors is withheld for intellectual property protection, but they are engineered to be robust, generalizable, and efficient to compute. Together with the scalar property and alignment, they provide a rich, low-dimensional representation suitable for a wide range of surface-sensitive prediction tasks.

### Architectural Considerations

AMPTCR representations require a .PDB file and a .PLY file to be generated. Once created, they are saved in compact .NPZ files, each containing a point cloud representing a molecule's surface manifold. Within this point cloud, each point contains its location in 3D space (after alignment using the heuristic described above), a scalar value representing the quantum property of the manifold at that location, and multiple channels representing topological information of the surrounding area.

For the purposes of assessing AMPTCR, a Dynamic Graph Convolutional Neural Network (DGCNN) was employed and adapted to best utilize the information contained in this novel molecular representation. The local K-Nearest Neighbor (K-NN) graph construction was limited to using each point's 3D positional data, while topology information is passed through a shared pointwise MLP to generate a fixed-dimensional embedding. To increase robustness to stochasticity in exact point positioning, a slight random jitter is applied to the position of each point whenever a data loader batch is retrieved during training. Optionally, a random rotation jitter can also be applied to the point cloud as a whole to increase robustness to minor errors in the alignment heuristic.

## Assessment of Performance

To assess the performance of the AMPTCR framework, two datasets were utilized. The first was a melting point dataset by [Bergstrom et al.](), consisting of 269 entries (after conversion), primarily small drug molecules. The second was a custom-curated dataset from ChEMBL of molecules with at least 4 recorded entries of inhibitory activity against *Escherichia coli*, with 50 or fewer atoms, a molecular weight less than 500, and a median molecular-weight-normalized inhibition value (µM) of 50 or less, totaling 521 entries after conversion.

### Alignment Heuristic

During development of the alignment heuristic, a series of rotation challenges were performed on diverse molecules to probe the limitations of the current technique and identify paths for improvement. The alignment procedure is generally robust, reliably orienting most molecules to nearly identical canonical frames despite random starting rotations. However, certain scenarios expose its shortcomings—particularly with highly symmetrical molecules, such as benzene derivatives with small substituents, where orientation may switch between multiple plausible alignments depending on the initial pose. [Fig. 1] illustrates a rotation challenge with the molecule valsartan: three initial orientations are shown in column [a], with their heuristic-aligned point clouds in column [b]. The top two align correctly to nearly identical frames, but the third demonstrates a sign flip along one principal axis, resulting in a mirrored configuration. While the current heuristic performs well overall, these occasional failures highlight the need for further refinement to fully resolve ambiguities and ensure consistent alignment across all molecular cases, putting less of the invariance burden on the model itself.

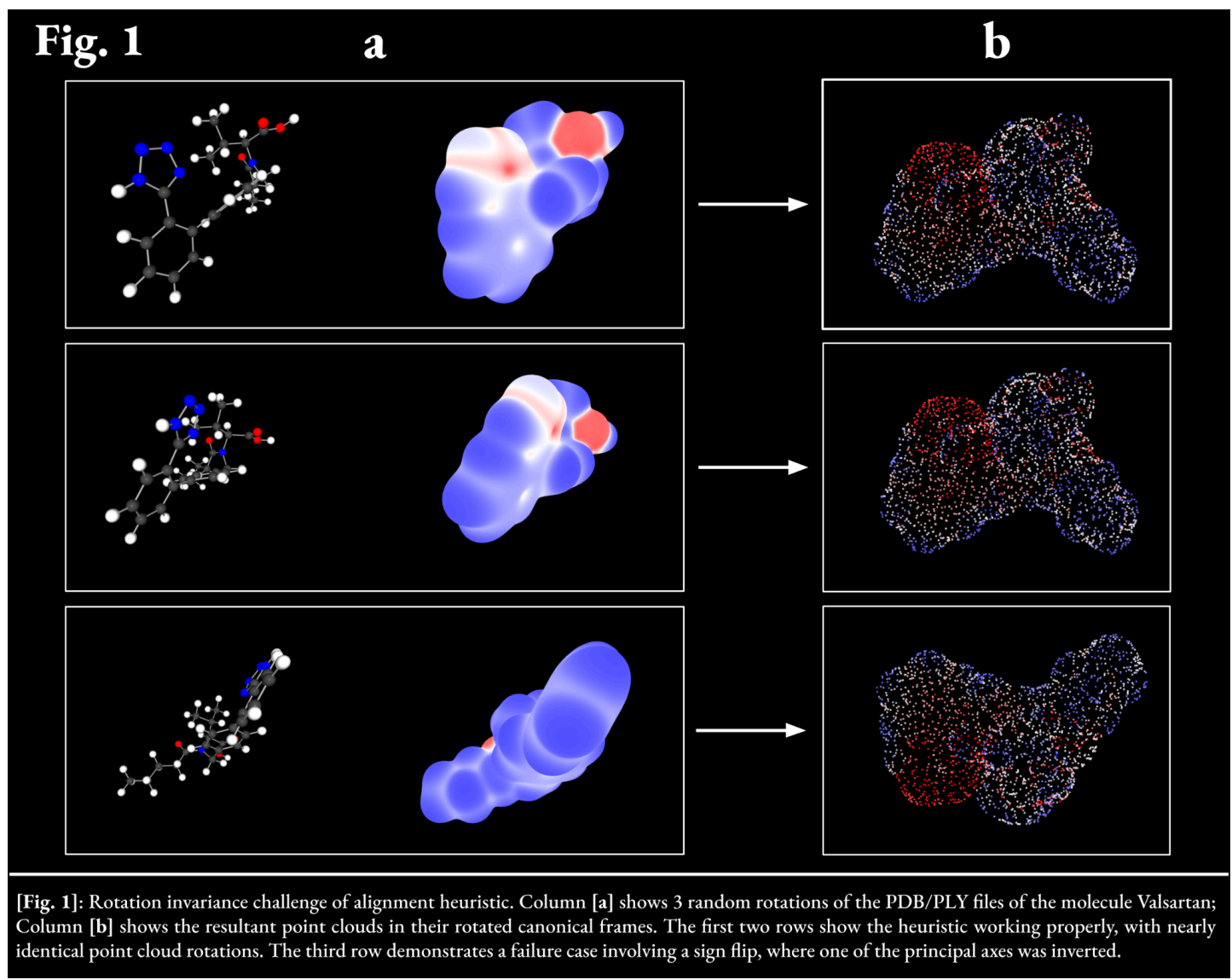

[Fig. 1]: Rotation invariance challenge of alignment heuristic. Column [a] shows 3 random rotations of the PDB/PLY files of the molecule Valsartan; Column [b] shows the resultant point clouds in their rotated canonical frames. The first two rows show the heuristic working properly, with nearly identical point cloud rotations. The third row demonstrates a failure case involving a sign flip, where one of the principal axes was inverted.

**Initial Deep Learning Performance**

First, the basic functionality of both AMPTCR and the DGCNN architecture was assessed by predicting the molecular weight of entries in the Bergstrom database, a trivial task to ensure basic information is actually encoded. This consisted of six-fold cross validation, where for each fold the data was randomly split into training and validation sets in a 95:5 ratio, after which the model was trained for exactly 30 epochs. Upon training completion, both training and validation datasets were input into the model, the predictions saved, and the model reset so the next fold could commence. Once all six folds were completed, the training and validation predictions across all folds were aggregated and analyzed. No early stopping, fold repeats, or best epoch selection occurred. For molecular weight predictions, a 256-point representation was used for each molecule—the smallest tested point number—and ESP was the quantum scaler of choice. Due to the small number of points, model K-NN was set to 1, meaning no graph functionality was active. The results of six folds are shown in [Fig. 2], with validation data achieving an $R^2$ of 0.82 and a regression slope of 0.88 [b]. Training data predictions, while numerically superior [a], were fairly comparable to validation predictions in terms of performance.

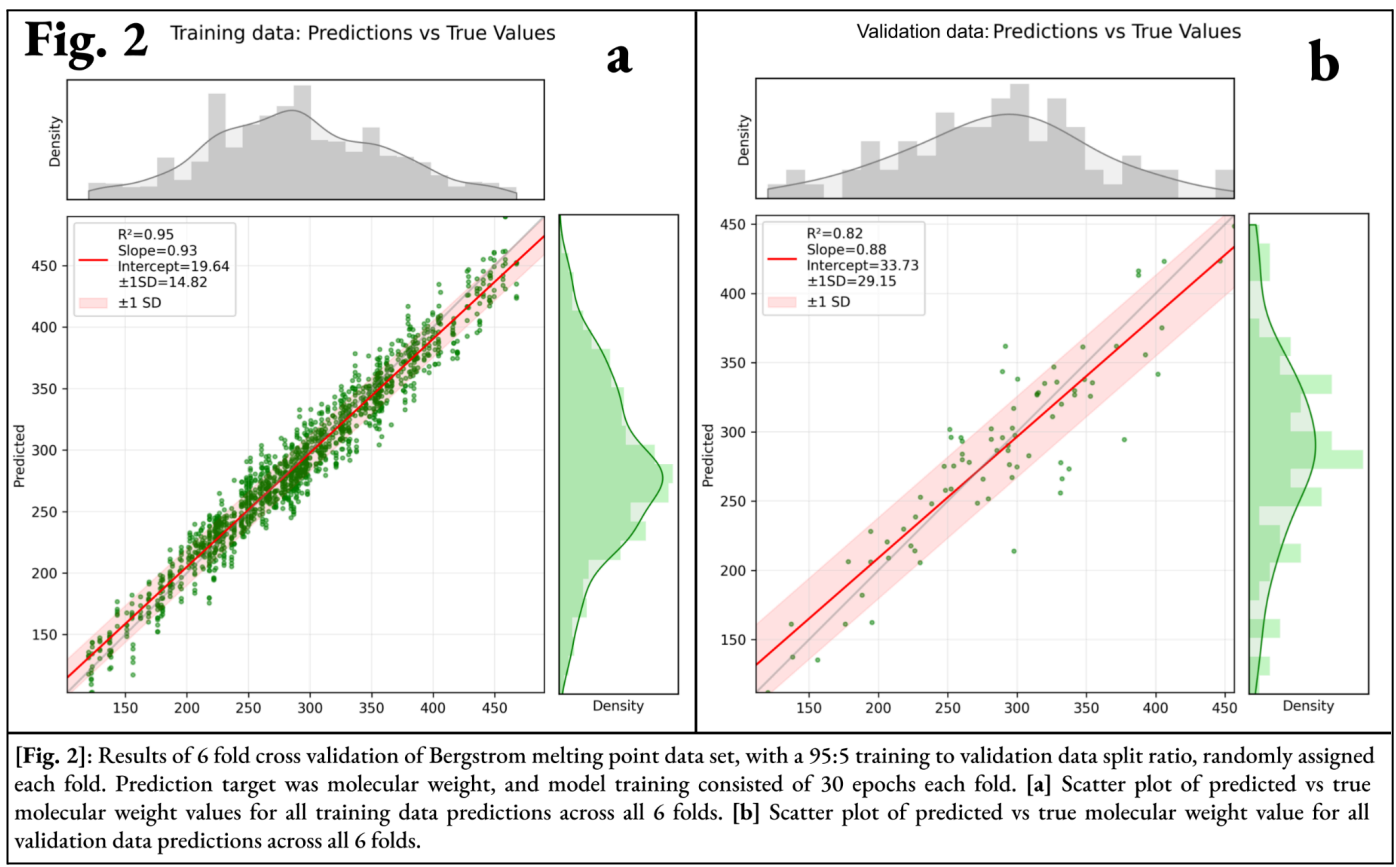

[**Fig. 2**]: Results of 6 fold cross validation of Bergstrom melting point data set, with a 95:5 training to validation data split ratio, randomly assigned each fold. Prediction target was molecular weight, and model training consisted of 30 epochs each fold. [a] Scatter plot of predicted vs true molecular weight values for all training data predictions across all 6 folds. [b] Scatter plot of predicted vs true molecular weight value for all validation data predictions across all 6 folds.

**Post Hoc Prediction Realignment**

In cases where meaningful patterns are learned, but the model is overly cautious with prediction magnitude, the following strategy was devised: After training on a given fold, a single affine transform is fitted to the training-set predictions to eliminate residual slope bias. Parameters p and q are determined by ordinary least squares by fitting $\hat{y} = py + q$ to the model's predictions $\hat{y}$ and true labels $y$.

The inverse mapping $y' = \dfrac{\hat{y} - q}{p}$ is then applied uniformly to both training and validation predictions.

This monotonic, one-to-one rescaling preserves the original ranking of molecules and depends solely on the training labels. Calibration performed per fold without reference to validation labels prevents any data leakage. Conceptually, this functions as a closed-form linear regression head appended post hoc, correcting multiplicative and additive biases without further gradient updates. In practice, the regression slope on the training set is restored to unity, residual variance remains unchanged, and the adjustment carries over to unseen data with negligible variance inflation, yielding calibrated predictions without violating the integrity of validation data, both within a fold and across all folds.

Part of the reason this post hoc affine calibration was employed was due to the number of training epochs being fairly limited before overfitting would occur. Oftentimes, these epochs would be enough for the model to learn meaningful signals and patterns within the data, but not enough to force the optimizer to properly scale predictions to their true magnitude, instead hugging the mean. This alignment allows these learned signals to be properly rescaled without requiring a higher epoch number or learning rate, thereby helping minimize overfitting.

**Molecular Weight Prediction with Post Hoc Realignment**

The molecular weight prediction assessment on the Bergstrom dataset was repeated using 1024 points per molecule and a K-NN value of 20, thereby enabling full graph-based model functionality. To ensure that validation data closely reflected the overall dataset distribution, a 90:10 train–validation split was used within each fold. A comprehensive 16-fold cross-validation was performed, training the network for exactly 20 epochs per fold. This entire process was conducted twice: first with raw model predictions, and then with post hoc realignment applied to predictions within each fold.

When using raw predictions, pronounced and consistent underprediction was observed in both training and validation sets, being an order of magnitude more severe than the previous run. Importantly, however, regression lines for both sets remained tightly grouped, with nearly identical slopes, indicating that the network was capturing meaningful structural information but systematically underpredicting value magnitude. Contrastingly, in the run which utilized post hoc realignment, regression performance improved substantially: across all 16 folds, cumulative validation data achieved a regression slope of 1.01 and an $R^2$ of 0.87 **[Fig. 3][b]**. Raw validation data predictions yielded a much lower slope of 0.25—virtually identical to that of the training data—while maintaining a similar $R^2$ of 0.85 **[a]**.

The underlying cause of this systemic underprediction remains uncertain, but is suspected to stem from the large absolute values in molecular weight prediction, coupled with the relatively limited number of training epochs. This combination likely resulted in the network capturing relevant features but compressing prediction magnitudes toward the mean. Notably, this underprediction effect was mostly absent in the 256-point, K-NN = 1 run, suggesting that increased data fidelity and activation of graph functionality may themselves contribute to the observed prediction compression.

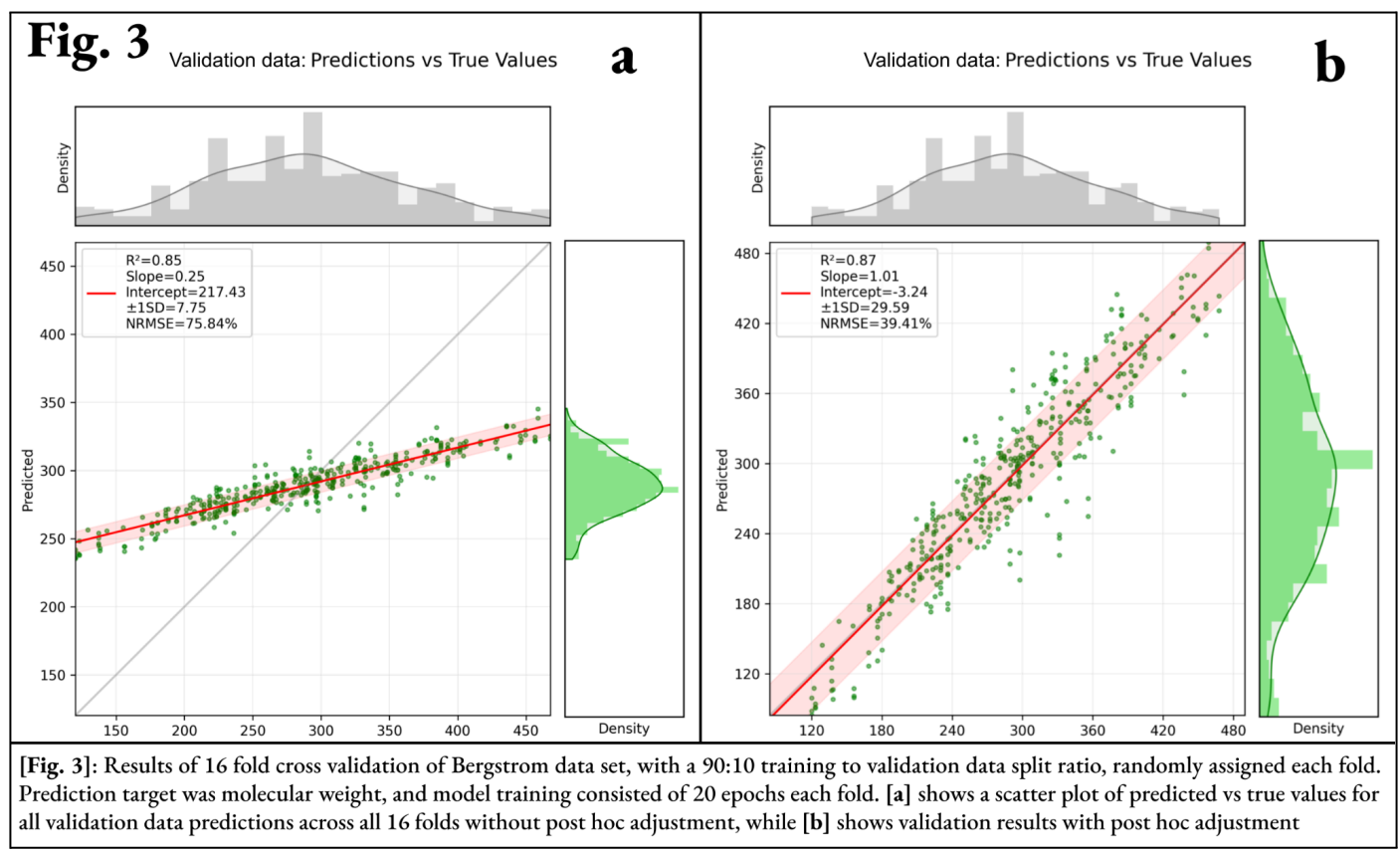

**[Fig. 3]**: Results of 16 fold cross validation of Bergstrom data set, with a 90:10 training to validation data split ratio, randomly assigned each fold. Prediction target was molecular weight, and model training consisted of 20 epochs each fold. **[a]** shows a scatter plot of predicted vs true values for all validation data predictions across all 16 folds without post hoc adjustment, while **[b]** shows validation results with post hoc adjustment

**Relational Attention Mechanism**

For the purposes of improving performance, especially regarding complex, non-localized properties, such as bacterial growth inhibition, a custom designed attention mechanism was developed to fully utilize the rich assortment of information encoded by AMPTCR. The attention module augments standard multi-head self-attention with three auxiliary, physically grounded relational bias channels—geometric, quantum, and topological—enabling the model to capture chemically relevant interactions between surface points.

Given per-point features $\mathbf{x} \in \mathbb{R}^{B \times F \times N}$, the input is linearly projected into queries, keys, and values and reshaped into $H$ attention heads of dimension $d = F/H$. For each head, attention logits are computed via scaled dot product $\langle \mathbf{q}_i^h, \mathbf{k}_j^h \rangle / \sqrt{d}$, forming a pairwise affinity tensor across all points.

Three additional bias tensors are computed and added to these affinities before the softmax. First, the geometric bias is derived from the 3D coordinate displacement $\Delta \mathbf{r}_{ij} = \mathbf{p}_i - \mathbf{p}_j$, projected via a shared $1 \times 1$ convolution across all heads. Second, the quantum bias is formed from the difference in scalar surface properties $q_i - q_j$, such as dual-Fukui indices or ESP, similarly convolved. Third, the topological bias is computed by taking the dot product of the first intrinsic topology vectors $\langle \mathbf{t}i, \mathbf{t}j \rangle$, capturing alignment between local surface frames. Each head maintains a learnable gating vector $\sigma^h \in (0,1)^3$, applied via sigmoid, which weights the contribution of each bias channel. The final attention logits are given by the sum of the scaled dot product and the gated biases:

$$\tilde{A}^h ij = \frac{\langle \mathbf{q}i^h, \mathbf{k}j^h \rangle}{\sqrt{d}} + \sigma_1^h G^h ij + \sigma_2^h E^h ij + \sigma_3^h T^h ij.$$

Following softmax normalization, the weighted values are aggregated and passed through a projection layer. The result is added to the input via a residual connection and passed through a feed-forward network with GELU activation, dropout, and a second residual path. Layer normalization is applied before and after the feed-forward block. This architecture enables the network to refine per-point features based not only on learned interactions, but also on interpretable geometric and quantum relationships, while maintaining invariance to global rotations and alignment heuristic sign flips.

This attention mechanism enhances the network's ability to differentiate and learn from the three distinct signals encoded at each surface point—normalized position, quantum scalar value, and topology vectors—by modeling their individual contributions to the molecule's overall property. It also reinforces separation between these signals, reducing the risk of their contributions being inadvertently conflated, which could otherwise impair model performance.

**Complex Deep Learning Assessment: Growth Inhibition**

  Having demonstrated that AMPTCR and the DGCNN framework could successfully encode and predict basic molecular properties, a more demanding and biologically relevant test was undertaken to evaluate the representation's utility in its intended domain. Given AMPTCR's capacity to capture both the intrinsic 3D manifold geometry of a molecule and its electronic property distribution across that surface, predicting bacterial growth inhibition was selected as the benchmark. In the context of antibiotic activity, both electronic and geometric molecular features are critical—manifold shape influences membrane permeability, receptor binding, and a host of other biological interactions that determine whether a compound inhibits or kills bacteria.

  For this evaluation, a curated subset of the ChEMBL database was used (n=521)[Fig. 4], containing small molecules with experimentally measured growth inhibition values against *E. coli* [a]. Rather than using electrostatic potential (ESP), Dual Fukui functions ($F_2$) were selected as the electronic descriptor during .PLY surface generation, due to their higher biological relevance in redox-sensitive systems. Each molecule was sampled at 1024 points across its surface, and Morgan Fingerprints were used as auxiliary training data, as is standard in the literature ([Zagidullin et al.](#)). The fingerprint vector is mapped to a scalar by a dedicated MLP, then linearly blended with the AMPTCR output using an FP weight; for example, FP weight = 0.2 yields 20% fingerprint contribution. An FP weight of 0.25 (binary) and 0.15 (regression) were used.

  For bactericity prediction, both binary classification and regression were performed. In most literature, compounds with inhibition values below 1 µM are considered highly antibacterial, while those above 10 µM are typically regarded as inactive ([Biswas et al.](#)). Using this as a framework, the ChEMBL dataset was binarized: compounds with a median MIC ≤ 1 µM were labeled as hits, and those with MIC > 10 µM as non-hits. Molecules with intermediate values were excluded to enforce clearer class separation, given the otherwise continuous distribution of inhibition values. This filtering yielded a dataset of 405 compounds—252 hits and 153 non-hits—resulting in a class imbalance ratio of approximately 1.65:1, and was used for binary classification. The full, n=521 dataset was used for regression, with a log10 transformation applied to normalize the distribution[Fig. 4][b].

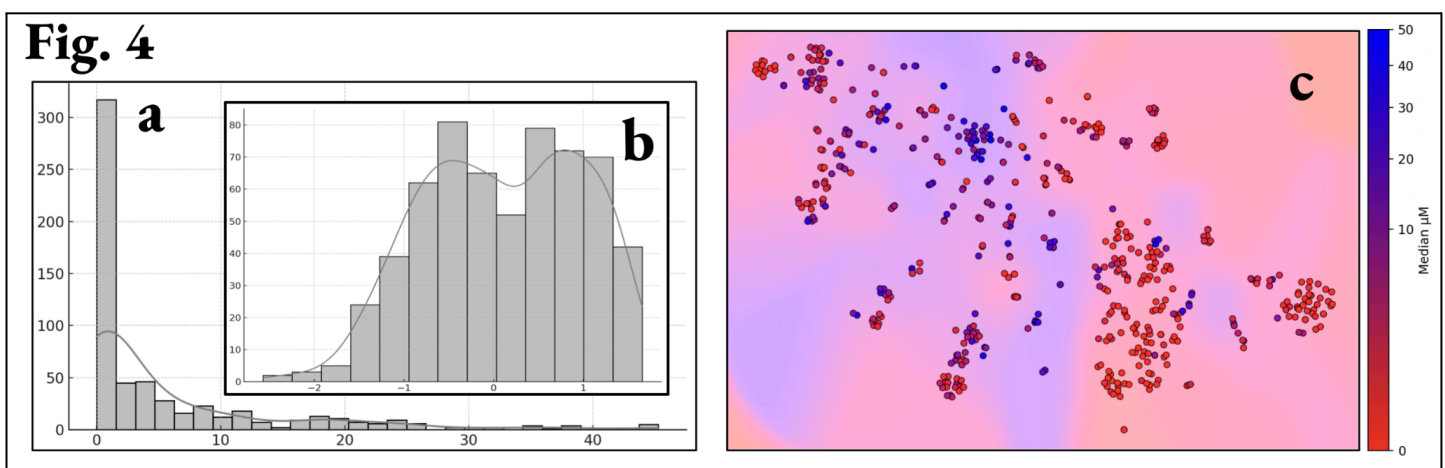

[Fig. 4]: Relevant information regarding custom curated E.coli inhibition dataset (n=521). [a] Distribution of mean, molecular-weight-normalized growth inhibition values (µM). [b] Distribution of growth inhibition values after log 10 transformation. [c] t-SNE projection of molecules based on Tanimoto similarity of Morgan fingerprints, with low µM values shown in red, and high µM values in blue (log scaled coloring).

**Binary Classification Results**

On the binarized dataset, a 6 fold cross-validation was performed, where for each fold the model was trained for 25 epochs, after which validation data would be input and predictions recorded. K-Fold splits were used, and no early stopping, fold repeats, or epoch selection was done. The results consisted of a mean ROC AUC of 0.912 (± 0.011 SE), a precision of 0.881, and a recall of 0.881 **[Fig. 5][a][b]**. Notable discrimination between the two classes is observed **[c]**. An identical run where only Morgan fingerprints were used yielded a significantly lower mean ROC AUC of 0.715 ± 0.021, showing AMPTCR contributed significantly to model performance.

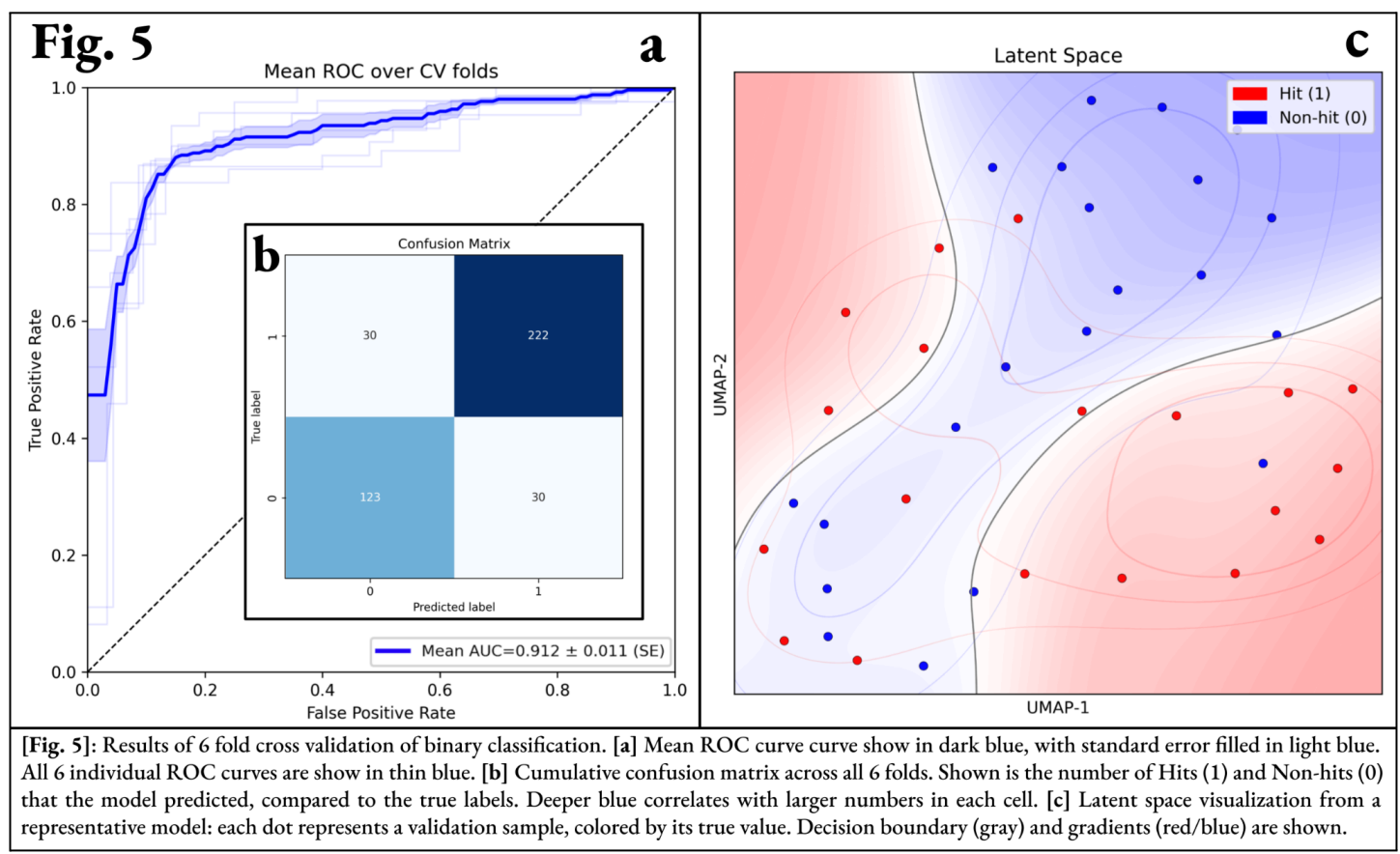

**[Fig. 5]**: Results of 6 fold cross validation of binary classification. **[a]** Mean ROC curve curve show in dark blue, with standard error filled in light blue. All 6 individual ROC curves are show in thin blue. **[b]** Cumulative confusion matrix across all 6 folds. Shown is the number of Hits (1) and Non-hits (0) that the model predicted, compared to the true labels. Deeper blue correlates with larger numbers in each cell. **[c]** Latent space visualization from a representative model: each dot represents a validation sample, colored by its true value. Decision boundary (gray) and gradients (red/blue) are shown.

**Regression Results**

For regression, the entire n=521 ChEMBL dataset was used, with a log10 transformation applied to the median μM value for each entry. To assess performance, a 24 fold cross-validation was conducted, with random splits (90:10 ratio) being used. During each fold, the model would train for 25 epochs, after which validation and training data would be input into the network. Training data predictions would undergo post hoc realignment, and the same transformation would be blindly applied to validation data. As this alignment would be applied every fold, no data leakage would occur and the integrity of the validation data would not be violated. No early stopping, fold repeats, or epoch selection occurred. Across the 24 fold cross-validation, the model achieved a mean regression slope of 0.71, an $R^2$ of 0.54, and a standard deviation of ± 0.58 log10 μM **[Fig. 6][a]**. While regression on inhibition values is relatively uncommon—typically supplanted by binary classification—these results are notable. When compared to existing literature, they are broadly equivalent to those reported by Bajiya et al., despite their use of a significantly larger dataset of 3,929 peptides. An identical run where only Morgan fingerprints were used yielded an inconsequential performance, comprising a mean regression slope of 0.53, an R² of 0.13, and a standard deviation of ± 1.22 log10 μM, again demonstrating the significant predictive value added by AMPTCR.

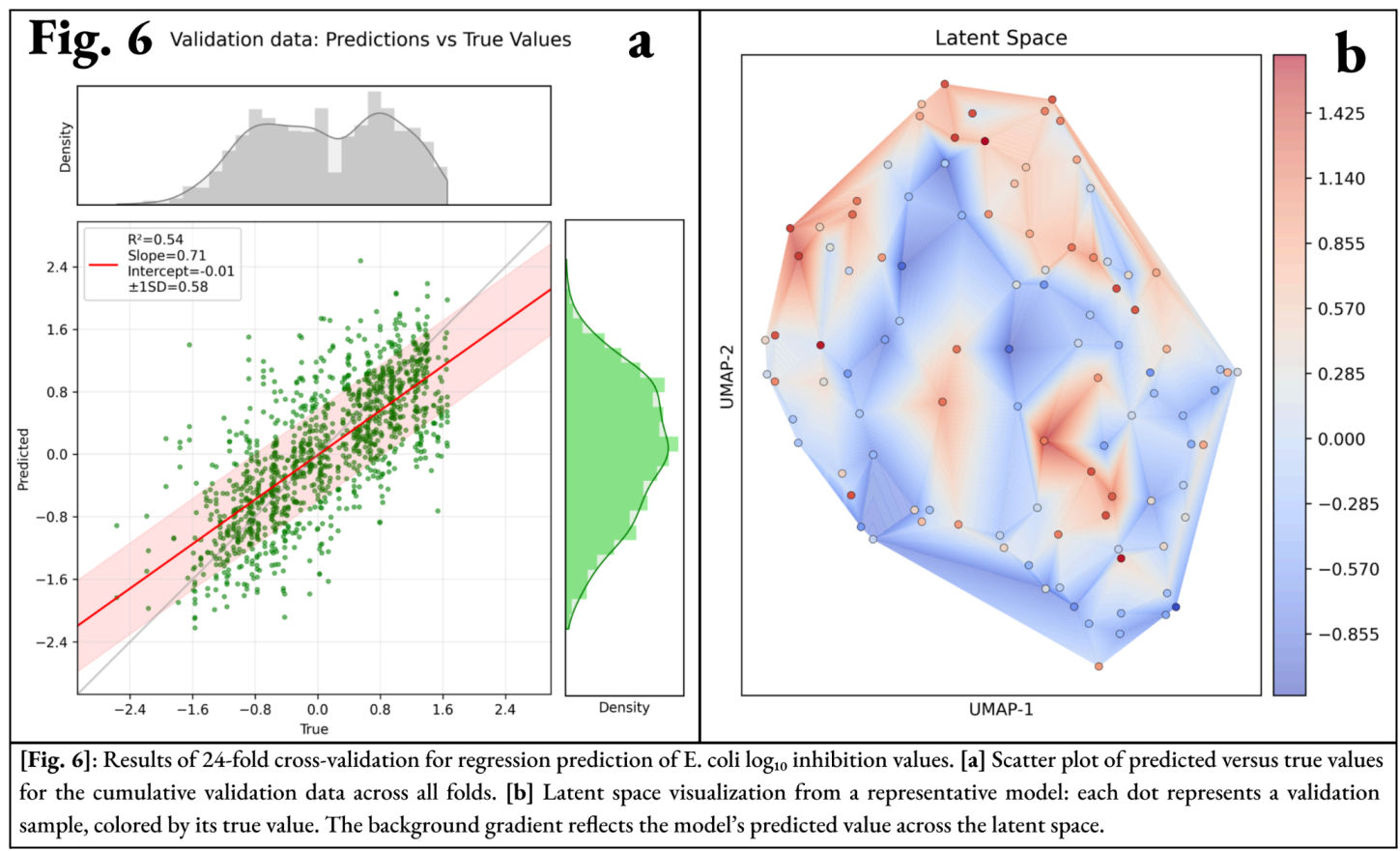

[Fig. 6]: Results of 24-fold cross-validation for regression prediction of E. coli $\log_{10}$ inhibition values. [a] Scatter plot of predicted versus true values for the cumulative validation data across all folds. [b] Latent space visualization from a representative model: each dot represents a validation sample, colored by its true value. The background gradient reflects the model's predicted value across the latent space.

## Discussion and Conclusion

This work introduces and assesses the capabilities of AMPTCR, a novel point cloud–based representation of molecular surfaces that integrates local quantum-derived scalar fields and geodesically informed topological descriptors within a canonical alignment framework. By shifting the burden of spatial invariance from the model to the data itself, AMPTCR enables accurate learning of surface-local properties using conventional SE(3)-sensitive neural networks, avoiding the substantial computational overhead typically associated with SE(3)-equivariant architectures. This design not only reduces model complexity but also enhances generality and interpretability, allowing more seamless integration with standard graph or point-based deep learning frameworks.

The results presented demonstrate that AMPTCR captures chemically and physically meaningful patterns on molecular surfaces. In the molecular weight prediction task—a baseline designed to validate the expressiveness of the representation—AMPTCR achieved strong predictive performance ($R^2 = 0.87$) following a simple post hoc calibration step, suggesting that the learned features maintain a robust linear relationship with ground-truth properties. Importantly, this performance remained consistent across representations of varying surface fidelity (from 256 to 1024 points) and under graph-based modeling conditions, supporting the method's robustness and flexibility. The observed attenuation in raw prediction slopes is a common challenge when regressing targets under limited training durations, but was effectively mitigated using fold-wise affine recalibration, applied without any data leakage occurring.

More critically, AMPTCR demonstrated strong performance in predicting bacterial growth inhibition—a biologically significant task for which this representation was specifically designed. In this context, surface-local quantum and geometric features are especially important, as they influence membrane permeability, binding affinity, and broader biological activity. Binary classification achieved high discriminative power (ROC AUC = 0.912), despite moderate class imbalance, indicating that AMPTCR encodes features predictive of antibiotic activity. This is further supported by the regression results on bacterial inhibition, which achieved performance comparable to studies using datasets more than seven times larger. Together, these findings position AMPTCR as a compelling alternative or supplement to traditional molecular representations that often overlook surface-mediated interactions, especially in domains such as antibiotic discovery.

AMPTCR is not intended to replace graph- or SMILES-based methods outright. Instead, it offers a complementary modality: one that encodes the physical interface through which molecules interact with their environments. Its compatibility with many standard neural architectures makes it a practical component in multi-modal learning strategies or ensemble pipelines, where it can serve as either a primary or auxiliary molecular representation, depending on the task—though so far, this has only been explored in combination with Morgan fingerprints.

Despite promising results, several areas remain open for exploration and refinement. The alignment heuristic, while broadly effective, can produce ambiguous outputs in highly symmetric molecules. Resolving these edge cases may require integration of orientation-resolving priors or symmetry-aware preprocessing. Although manual rotation challenges were used to probe heuristic robustness, a more systematic evaluation of alignment errors—such as the frequency of sign flips or average rotational deviations—would help benchmark the heuristic's consistency and limitations.

Similarly, while the current topological encoding appears to effectively capture curvature and surface shape features, its proprietary structure has not been exhaustively benchmarked against other descriptors, such as SHAP values or Zernike moments. Future work comparing these alternatives would clarify which aspects of local geometry most contribute to predictive accuracy. Additionally, exploring how variations in the topological encoding scheme—such as the density of encoded topological information per point and the geodesic extent over which features are computed—influence performance could help identify task-specific optimal configurations.

AMPTCR represents a lightweight yet expressive representation for encoding chemically meaningful surface information in small molecules. It supports accurate, data-efficient learning using conventional neural network architectures, without requiring full SE(3) equivariance, and shows promise in tasks where surface-level effects govern molecular behavior, such as those relevant in biological systems, and thereby pharmaceutical applications. Future work will focus on further benchmarking, expanding biological use cases, and integrating AMPTCR into multi-modal frameworks to enhance generalization and utility across drug discovery applications.

# References

[1] Li, T., Huls, N. J., Lu, S., & others. (2024). Unsupervised manifold embedding to encode molecular quantum information for supervised learning of chemical data. Communications Chemistry, 7, Article 133. https://doi.org/10.1038/s42004-024-01217-z

[2] Liu, R., Lauze, F., Bekkers, E. J., Darkner, S., & Erleben, K. (2025). SE(3) group convolutional neural networks and a study on group convolutions and equivariance for DWI segmentation. Frontiers in Artificial Intelligence, 8, Article 1369717. https://doi.org/10.3389/frai.2025.1369717

[3] Bergström, C. A. S., Norinder, U., Luthman, K., & Artursson, P. (2003). Molecular descriptors influencing melting point and their role in classification of solid drugs. Journal of Chemical Information and Computer Sciences, 43(4), 1177–1185. https://doi.org/10.1021/ci020280x

[4] Zagidullin, B., Wang, Z., Guan, Y., Pitkänen, E., & Tang, J. (2021). Comparative analysis of molecular fingerprints in prediction of drug combination effects. Briefings in Bioinformatics, 22. https://doi.org/10.1093/bib/bbab291

[5] Biswas, D., Benson, S., Matunis, A., Gebretsadik, G., Wertz, A., StPierre, B. J., Schacht, N., Yan, Y., Gebremichael, H. Y., Wong, P. K., Baughn, A. D., & Medina, S. H. (2025). Lead informed artificial intelligence mining of antitubercular host defense peptides. Biomacromolecules, 26(5), 3167–3179. https://doi.org/10.1021/acs.biomac.5c00244

[6] Bajiya, N., Kumar, N., & Raghava, G. P. S. (2024). Prediction of inhibitory peptides against E. coli with desired MIC value. bioRxiv. https://doi.org/10.1101/2024.07.18.604028